\documentclass{article}

\usepackage[utf8]{inputenc}
\usepackage{amsmath}
\usepackage{amssymb}
\usepackage{amsthm}
\usepackage{graphicx}
\usepackage{natbib}
\usepackage{url}
\usepackage{hyperref}

\title{Learning in two-player games between transparent opponents}
\author{Adrian Hutter\\Google, Switzerland\\\texttt{adrianhutter@google.com}}
\date{\today}

\begin{document}

\maketitle

\begin{abstract}

We consider a scenario in which two reinforcement learning agents repeatedly play a matrix game against each other and update their parameters after each round.
The agents' decision-making is transparent to each other,  which allows each agent to predict how their opponent will play against them. To prevent an infinite regress of both agents recursively predicting each other indefinitely, each agent is required to give an opponent-independent response with some probability at least $\varepsilon$. 
Transparency also allows each agent to anticipate and shape the other agent's gradient step, i.e.\ to move to regions of parameter space in which the opponent's gradient points in a direction favourable to them. 
We study the resulting dynamics experimentally, using two algorithms from previous literature (LOLA and SOS) for opponent-aware learning.
We find that the combination of mutually transparent decision-making and opponent-aware learning robustly leads to mutual cooperation in a single-shot prisoner's dilemma.
In a game of chicken, in which both agents try to manoeuvre their opponent towards their preferred equilibrium, converging to a mutually beneficial outcome turns out to be much harder, and opponent-aware learning can even lead to worst-case outcomes for both agents. 
This highlights the need to develop opponent-aware learning algorithms that achieve acceptable outcomes in social dilemmas involving an equilibrium selection problem.

\end{abstract}

\section{Introduction}

Reinforcement learning is concerned with constructing agents that learn to achieve high rewards in a diverse set of environments. Multi-agent reinforcement learning studies learning in environments that contain other agents, possibly pursuing conflicting goals.
The simplest way do deal with the presence of other learning agents during learning is to treat them as just another part of the environment.
However, such an environment will be non-static and non-Markovian \citep{laurent11,lanctot17}.
The simplistic treatment also ignores the incentives that our agent's parameters and their updates create for the learning process of the other agents.
This is particularly relevant in so-called social dilemmas, in which agents can benefit from mutual cooperation but have an incentive to unilaterally defect \citep{leibo17}. A number of authors recently developed strategies that allow reinforcement learners to  reach mutual cooperation in social dilemmas \citep{lerer17,peysakhovich17,hughes18,wang18,foerster18,letcher18,baumann18,clifton20}.

The prisoner's dilemma (PD) is the canonical example of a social dilemma. Mutual cooperation in the PD is Pareto-superior to mutual defection, but defecting is a dominant strategy and mutual defection is the only Nash equilibrium of the game. Strategies like tit-for-tat enable mutual cooperation in the iterated PD \citep{axelrod81,harper17}, while cooperation is much more elusive in the single-shot case. 
Vanilla gradient descent learners fail to learn tit-for-tat and end up defecting with each other even when repeatedly interacting in an iterated PD.
One strand of recent work considered learning agents that are aware of each other's learning process, and use this to anticipate and shape each other's parameter updates \citep{foerster18,letcher18}. When repeatedly playing an iterated PD, such learners can learn strategies similar to tit-for-tat, leading to mutual cooperation. In this work, we shall take the idea of mutual awareness of the two learning agents one step further, as we will now discuss.

Outside of machine learning, different authors have discussed conditions under which agents might cooperate with each other even in a \emph{single-shot} PD. What connects all of these ideas is a condition which we might call \emph{mutual transparency}, i.e.\ agents having some insight into each other's inner workings.
\citet{hofstadter83} coined the term ``superrationality'' to describe a type of player that will cooperate with other superrational players, since they reason that both players will arrive at the same decision, and hence mutual defection and mutual cooperation are the only two consistent outcomes. \citet{mcafee84}, \citet{howard88}, and \citet{tennenholtz04} introduced a program equilibrium formalism, in which players do not directly chose actions in the game, but rather submit programs which chose actions and are given access to each other’s source code. Mutual cooperation is enabled by submitting a program that cooperates if and only if the opponent program is syntactically exactly identical. However, cooperation is rather fragile if it relies on programs being exact copies of each other, and so different authors have worked on making cooperation in a single-shot PD more robust when assuming mutual transparency of the agents \citep{hoek13,lavictoire14,critch19,oesterheld19}. These references rely on logical as opposed to syntactical properties of the agents to achieve cooperation.
\citet{critch19} considers agents that search through all possible proofs up to length $k$ to prove that their opponent will cooperate with them; if they find such a proof, they cooperate, otherwise they defect. Surprisingly, such agents cooperate with each other for large enough but finite $k$.
\citet{oesterheld19} introduces an agent called $\varepsilon\texttt{GroundedFairBot}$ which cooperates unconditionally with small probability $\varepsilon$, and otherwise mirrors the opponent's action when predicting how the opponent will act when playing against $\varepsilon\texttt{GroundedFairBot}$.
When facing a logically similar agent, the mutual recursive function calls terminate after a finite number $O(1/\varepsilon)$ with unit probability and cooperation ensues. When facing a defector, $\varepsilon\texttt{GroundedFairBot}$ defects with probability $1-\varepsilon$.

In this work, we bring together the idea of mutual transparency enabling cooperation with opponent-aware learning as developed in \citet{foerster18,letcher18}.
Similar to \citet{oesterheld19}, we allow both agents to predict how their opponent will play against them, while requiring them to give an opponent-independent response with probability at least $\varepsilon$. The agents can repeatedly interact in a social dilemma and can update their parameters after each round. 
An interesting question is whether policies similar to $\varepsilon\texttt{GroundedFairBot}$ can emerge as the result of such a learning process, which we shall answer in the affirmative. 

Opponent-aware learning is necessary to learn such mutually cooperating policies; simple gradient learners will always learn mutual defection.
We use two algorithms developed in previous literature for this, LOLA \citep{foerster18} and SOS \citep{letcher18}. We will find that both of these have their own (sometimes unexpected) advantages and drawbacks.

In addition to the PD, we consider another well-known social dilemma, the game of chicken. A key difference of the game of chicken from the PD is that it has multiple Nash equilibria, and both players prefer a different one. 
We find that unlike in the PD, opponent-aware learning can be used in the game of chicken to outmaneuver the opponent and navigate them towards one's preferred equilibrium. 
However, both players attempting to do this can result in a worst-case outcome of both players going straight.

We study this scenario from the perspective of the principals who deploy the learning agents to interact with each other. 
From the principals' perspective, we can regard this as a game in which a move corresponds to submitting a certain learning agent. An interesting question will thus be whether in the game which the principals are playing, a socially optimal outcome is easier to achieve than in the underlying game which the agents are playing (PD or chicken). We find that this is the case for the PD, but not for the game of chicken.

The rest of this work is organized as follows. 
Sec.~\ref{sec:lola} briefly introduces learning with opponent-learning awareness, as developed in recent literature, and adapt it to our framework. For illustration, we apply it to two simple two-player games.
Sec.~\ref{sec:simulation} formally discusses games in which both players have the ability to predict how their opponent will play against themselves.
Sec.~\ref{sec:dilemmas} contains our main results, in which we apply the techniques from Secs.~\ref{sec:lola} and \ref{sec:simulation} to two well-known social dilemmas, the prisoner's dilemma and the game of chicken.
Sec.~\ref{sec:discussion} contains our concluding remarks.

\section{Learning with opponent-learning awareness}\label{sec:lola}

The theory of learning while taking the opponent's learning into account as used in this work was developed by \citet{zhang10,foerster18,letcher18}. We briefly recapitulate these ideas and adapt them to our scenario.

Consider a ``game'' between two players $A$ and $B$, in which expected payoffs as a function of players' parameters are given by $V_A(\theta_A,\theta_B)$ and $V_B(\theta_A,\theta_B)$, respectively. In each round, player $A$ calculates a gradient of $V_A$ and updates their parameters in the direction of this gradient using a learning rate $\delta_A$. Let $\bar{\theta}_A$ and $\bar{\theta}_B$ denote players' current parameters. The ``naive'' gradient for player $A$ is then given by 
\begin{align}
\xi_{\text{naive},A}&=\nabla_{\theta_A}V_A(\theta_A,\theta_B)\vert_{\theta_A=\bar{\theta}_A,\theta_B=\bar{\theta}_B} \nonumber\\ &= \nabla_{\theta_A}V_A(\bar{\theta}_A,\bar{\theta}_B)\ . 
\end{align}
This ``naive'' gradient however implicitly assumes $\theta_B$ to be static and so ignores that $B$ is learning as well.

A more sophisticated way of calculating $A$'s gradient might take $B$'s learning into account. There are two different ways in which $B$'s learning affects the direction in which $A$'s gradient points: $A$ might want to calculate their gradient at parameters $\theta_B$ that are already updated by a gradient step of $B$ (in other words, \emph{anticipate} $B$'s gradient step); and $A$ might want to move towards regions of parameter space in which $B$'s gradient points into directions that benefit $A$ (in other words, \emph{shape} $B$'s gradient step). 

In order to arrive at this formally, $A$ considers a gradient step of $B$ in which $B$ updates their parameters using a learning rate $\eta_A$. 
The learning rate $\eta_A$, used by $A$ in order to anticipate and shape $B$'s learning, need not be identical to $B$'s actual learning rate $\delta_B$.
The LOLA (Learning with Opponent-Learning Awareness \citep{foerster18}) gradient of $A$ is then given by
\begin{align}
\xi_{\text{LOLA},A}
&=\nabla_{\theta_A}V_A(\theta_A,\theta_B + \eta_A\nabla_{\theta_B}V_B(\theta_A,\theta_B))\vert_{\theta_A=\bar{\theta}_A,\theta_B=\bar{\theta}_B} \nonumber\\
&=\nabla_{\theta_A}V_A(\bar{\theta}_A,\bar{\theta}_B + \eta_A\nabla_{\theta_B}V_B(\bar{\theta}_A,\bar{\theta}_B)) \nonumber\\
&\quad+\eta_A\nabla_{\theta_B}V_A(\bar{\theta}_A,\bar{\theta}_B + \eta_A\nabla_{\theta_B}V_B(\bar{\theta}_A,\bar{\theta}_B))
\nabla_{\theta_A}\nabla_{\theta_B}V_B(\bar{\theta}_A,\bar{\theta}_B) \label{eq:lola_exact} \\
&=\xi_{\text{naive},A} + \underbrace{\eta_A\nabla_{\theta_B}V_B(\bar{\theta}_A,\bar{\theta}_B)
\nabla_{\theta_B}\nabla_{\theta_A}V_A(\bar{\theta}_A,\bar{\theta}_B)}_{\text{``anticipate''}}\nonumber\\
&\quad+\underbrace{\eta_A\nabla_{\theta_B}V_A(\bar{\theta}_A,\bar{\theta}_B)
\nabla_{\theta_A}\nabla_{\theta_B}V_B(\bar{\theta}_A,\bar{\theta}_B)}_{\text{``shape''}}+O(\eta_A^2)\ . \label{eq:lola_1st_order}
\end{align}

\citet{zhang10} study the first summand in Eq.~(\ref{eq:lola_exact}), which (following \citet{letcher18}) we will call Look Ahead (LA). They prove that for two-action two-player games, LA leads to convergence to a Nash equilibrium for sufficiently small learning rates. \cite{foerster18} study the combination of the first and third summand in Eq.~(\ref{eq:lola_1st_order}), that is, the naive gradient and the leading order opponent-shaping correction. Among other results, they find that it leads to mutual cooperation in an iterated PD. The version of LOLA we will use in this work is given by Eq.~(\ref{eq:lola_exact}), so in contrast to \citet{foerster18} we also incorporate the effects of anticipating the opponent's gradient step, and do not perform a leading-order Taylor expansion.

\citet{letcher18} introduce SOS (Stable Opponent Shaping), which, when adapted to our framework, uses a gradient
\begin{align}
\xi_{\text{SOS},A}
&=\nabla_{\theta_A}V_A(\bar{\theta}_A,\bar{\theta}_B + \eta_A\nabla_{\theta_B}V_B(\bar{\theta}_A,\bar{\theta}_B)) \nonumber\\
&\quad+p\eta_A\nabla_{\theta_B}V_A(\bar{\theta}_A,\bar{\theta}_B + \eta_A\nabla_{\theta_B}V_B(\bar{\theta}_A,\bar{\theta}_B))
\nabla_{\theta_A}\nabla_{\theta_B}V_B(\bar{\theta}_A,\bar{\theta}_B) \label{eq:sos_exact} \\
&=\xi_{\text{naive},A}+\eta_A\nabla_{\theta_B}V_B(\bar{\theta}_A,\bar{\theta}_B)
\nabla_{\theta_B}\nabla_{\theta_A}V_A(\bar{\theta}_A,\bar{\theta}_B) \nonumber\\
&\quad+p\eta_A\nabla_{\theta_B}V_A(\bar{\theta}_A,\bar{\theta}_B)
\nabla_{\theta_A}\nabla_{\theta_B}V_B(\bar{\theta}_A,\bar{\theta}_B)+O(\eta_A^2)\ . \label{eq:sos_1st_order}
\end{align}
These are identical to Eqs.~(\ref{eq:lola_exact}) and (\ref{eq:lola_1st_order}), respectively, up to a factor $p\in[0,1]$ which is used to tune the strength of the shaping term and is re-calculated for each gradient step. \citet{letcher18} prove that SOS converges to stable fixed points (SFPs) in a broad class of games and demonstrate that two SOS learners can avoid certain cooperation failures that two LOLA learners fall victim to; see Sec.~\ref{sec:tandem} for further discussion of this point. The scaling factor $p$ is thereby chosen as large as possible while guaranteeing that a) $\xi_{\text{SOS},A}$ always has positive inner product with the $p=0$ version (i.e., the LA-only version); and b) $\xi_{\text{SOS},A}$ goes to zero whenever $\xi_{\text{naive},A}$ does. 
While \citet{letcher18} use Eq.~(\ref{eq:sos_1st_order}), we will again avoid the leading-order Taylor expansion, and use the exact Eq.~(\ref{eq:sos_exact}).

In \citet{letcher18}, the factor $p$ is at each step identical for all involved learners. This implicitly assumes a certain form of global coordination which we cannot assume for our purposes. We will thus use a form of SOS in which each player calculates $p$ independently of all others, as proposed in Remark~4.7 in \citet{letcher18Thesis}. As shown in this remark, this version of SOS inherits convergence guarantees from the version used in \citet{letcher18}. The detailed calculation of $p$ is described in Appendix~\ref{app:sos}.

As discussed, there are some technical differences between the way LOLA and SOS gradients are calculated in \citet{foerster18} and \citet{letcher18}, respectively, and the way they are calculated in this work.
A conceptually more interesting difference is that we do not assume $\eta_A$ and $\eta_B$ to be fixed and equal to each other. We consider the choice of $\eta$ (as well as the choice between LOLA and SOS) to be part of the strategy of the principal who deploys a learning agent, and study the effects of choosing different values for $\eta$.
For instance, $\eta_A\rightarrow0$ is equivalent to calculating the naive gradient, while choosing $\eta_A$ much larger than $\delta_B$ corresponds to $A$ looking multiple steps of $B$ ahead. 
Larger values of $\eta_A$ and $\eta_B$ might be particularly relevant in competitive scenarios, in which both players attempt to outwit each other.

For illustration, we study the effects of LOLA and SOS on two simple games that do not yet involve predicting the opponent's action against oneself.

\subsection{Ultimatum game}

We consider a binary version of the ultimatum game \citep{guth82,sanfey03,henrich04,oosterbeek04} in which player $A$ (the proposer) receives a pot of \$10 and can choose between a ``fair'' split (\$5 each for them and their opponent) or an ``unfair'' split (\$8 for themselves, \$2 for the opponent). Player $B$ (the responder) can choose to accept the proposed split, or reject it, in which case both players receive nothing. We assume that $B$ will always accept fair splits, so both players have a single parameter describing their strategy. Player $A$'s parameter $\theta_A$ describes their probability of proposing a fair split, $p_\text{fair}=\sigma(\theta_A)$, where $\sigma$ denotes the sigmoid function. Player $B$'s parameter describes the probability of accepting a proposed unfair split, $p_\text{accept}=\sigma(\theta_B)$. Payoffs are given by
\begin{align}
V_A &= 5\,p_\text{fair} + 8\,(1-p_\text{fair})\,p_\text{accept} \nonumber\\
V_B &= 5\,p_\text{fair} + 2\,(1-p_\text{fair})\,p_\text{accept}\ , \nonumber\\
\end{align}
and so the naive gradients are
\begin{align}
\xi_{\text{naive},A} &= \frac{\partial V_A}{\partial p_\text{fair}}\frac{\partial p_\text{fair}}{\partial\theta_A} = (5 - 8\,p_\text{accept})\frac{\partial p_\text{fair}}{\partial\theta_A} \nonumber\\
\xi_{\text{naive},B} &= \frac{\partial V_B}{\partial p_\text{accept}}\frac{\partial p_\text{accept}}{\partial\theta_B} 
= 2\,(1-p_\text{fair})\frac{\partial p_\text{accept}}{\partial\theta_B} \ .
\end{align}
Player $B$'s naive gradient is thus always positive (since $p_\text{fair}<1$ for any finite $\theta_A$), while $A$'s naive gradient is positive when $p_\text{accept}\leq\frac{5}{8}$ and negative otherwise. 

\begin{figure}\centering
\makebox[\textwidth]{
\noindent\begin{tabular}{ccc}
      \includegraphics[width=40mm]{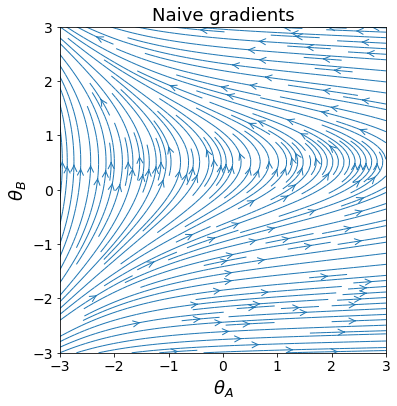} & 
      \includegraphics[width=40mm]{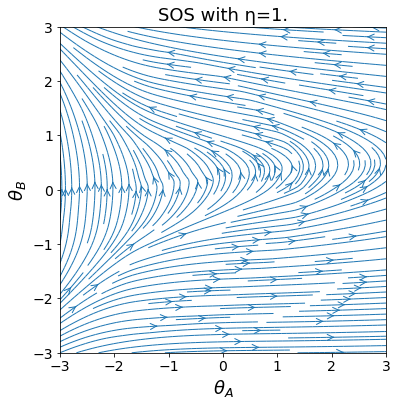} & 
      \includegraphics[width=40mm]{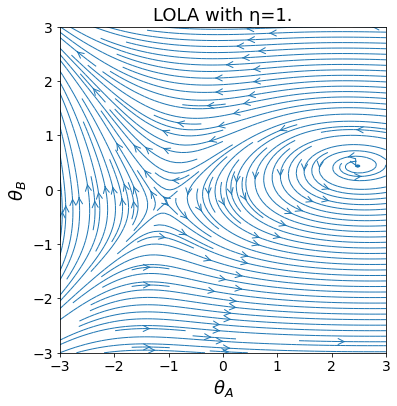} 
\end{tabular}
}
\caption{Gradient fields in the ultimatum game for naive learners (left), SOS learners (center), and LOLA learners (right) using $\eta_A=\eta_B=1$.}
\label{fig:ug}
\end{figure}

The full gradient fields for naive learners, SOS learners, and LOLA learners (with $\eta_A=\eta_B=1$) are shown in Fig.~\ref{fig:ug}. 
The field for SOS learners looks very similar to that for naive learners.
In particular, player $B$'s gradient is always positive, which is expected:
Since player $B$'s naive gradient is always positive, so is their LA gradient; the SOS gradient always has positive inner product with the LA gradient, meaning for agents with a single parameter that they always have the same sign. So the SOS gradient of player $B$ cannot become negative.

Things change when using LOLA with sufficiently high $\eta_B=1$: $B$'s gradient now becomes negative in certain regions of parameter space, meaning that the responder decreases their likelihood of accepting an unfair proposal, since they take into account that a low $p_\text{accept}$ will provide an incentive for the proposer to increase $p_\text{fair}$.

\subsection{Tandem game}\label{sec:tandem}

\citet{letcher18} introduce the tandem game as an example of a game in which LOLA learners' attempt at shaping each other's gradients leads to worse outcomes for both of them, which is avoided by SOS learners. In this game, both players have a single parameter, $\theta_A=x$ and $\theta_B=y$, and payoffs are given by
\begin{align}
V_A &= -(x+y)^2 + 2x \nonumber\\
V_B &= -(x+y)^2 + 2y\ .
\end{align}
The first summand in the payoff functions encourages the players to coordinate their choice of parameters to satisfy $x\approx-y$, while the second summand encourages each player to choose their parameter as large as possible.

The overall welfare can be written as
\begin{align}
V_A+V_B = -2(x+y-\frac{1}{2})^2+\frac{1}{8}\ .
\end{align}
When both players use LOLA gradients as in Eq.~(\ref{eq:lola_exact}) with $\eta=\eta_A=\eta_B$, straightforward calculus shows that the set of SFPs is described by 
\begin{align}
x+y = \frac{1-2\eta+4\eta^2}{(1-2\eta)^2}\ ,
\end{align}
which increases monotonically as a function of $\eta\in[0,\frac{1}{2})$ and diverges as $\eta\rightarrow\frac{1}{2}^{-}$.
The resulting overall welfare $V_A+V_B$ thus decreases monotonically as a function of $\eta\in[0,\frac{1}{2})$. 

LOLA learners display ``arrogant behavior'' \citep{letcher18} -- LOLA encourages both players 
to increase their own parameter, expecting that this will compel (via the first summand in the payoff) their opponent to decrease their own. SOS by contrast preserves the SFPs described by $x+y=1$ of two naive learners, leading to higher overall welfare.

\begin{figure}\centering
\includegraphics[width=1.0\textwidth]{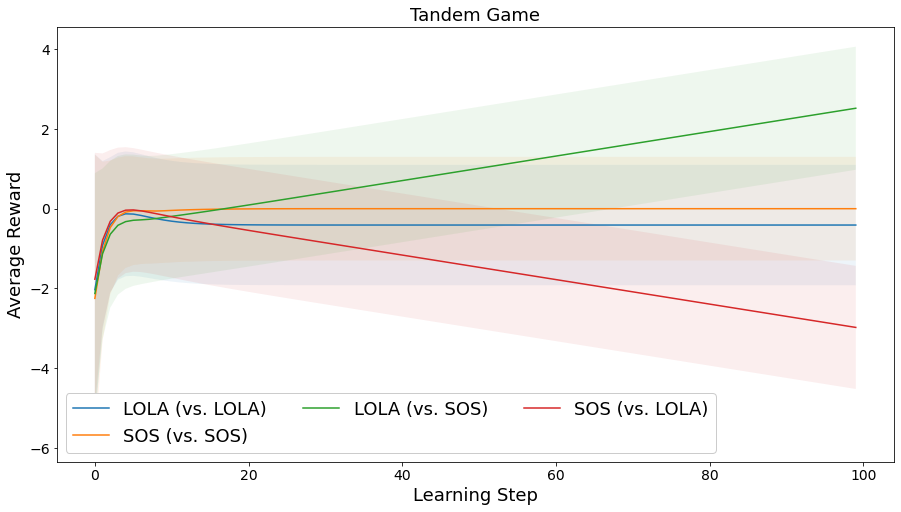}
\caption{Average rewards for different learners (LOLA or SOS) when playing the tandem game against other learners. Results are sampled over $100$ runs, shaded regions show one standard deviation.}
\label{fig:tandem}
\end{figure}

In the following, we simulate the learning dynamics of LOLA and SOS learners not only against themselves, but also against each other. We use $\delta_A=\delta_B=\eta_A=\eta_B=0.1$ and draw initial parameters from $\mathcal{N}(0, 1)$. Results are in Fig.~(\ref{fig:tandem}). While two SOS learners against each other receive higher rewards on average than two LOLA learners against each other, a LOLA learner is able to exploit an SOS learner.

If we consider the ``game'' in which two principals have to choose a learning algorithm for their agents, the choice between LOLA and SOS thus turns into a prisoner's dilemma: both players choosing SOS is better for both players than both players choosing LOLA, but choosing LOLA is a dominant strategy (i.e., leads to better outcomes than choosing SOS irrespective of the opponent's choice). 
Without some coordination regarding the choice of learning algorithm, we should thus expect that LOLA will be chosen in a competitive scenario.

\section{Decision-making with opponent transparency}\label{sec:simulation}

Consider a normal form game between two players $A$ and $B$. We assume that both players' decision making is transparent to each other, and so each player is able to predict their opponent's action against an other player, including themselves. Assume that players $A$ and $B$ have $d_A$ and $d_B$ actions available, respectively. The probabilistic reaction of $A$ to predicted actions of $B$ (when predicting how $B$ will play against $A$) can be summarized in a $d_A\times d_B$ matrix $M_A\in\mathbb{R}_{\geq0}^{d_A\times d_B}$, with non-negative entries and columns that sum to unity. We further denote $A$'s probability of choosing an action based on predicting $B$ (as opposed to an opponent-independent action) by $p_{P,A}$, and their ($d_A$-dimensional) probability distribution over actions when choosing their action independently of their opponent by $v_A\in\mathbb{R}_{\geq0}^{d_A}$. Player $A$'s policy is thus completely determined by the tuple $(p_{P,A}, v_A, M_A)$, and analogously for player $B$. 
$\varepsilon\texttt{GroundedFairBot}$ introduced by \citet{oesterheld19} is an example of an agent fitting into this general framework.

Clearly, having both $p_{P,A}\rightarrow1$ and $p_{P,B}\rightarrow1$ will lead to an infinite regress, so we require that both $p_{P,A}\leq1-\varepsilon$ and $p_{P,B}\leq1-\varepsilon$, where in the following we will use $\varepsilon=10^{-3}$. 
The resulting probability distribution $w_A\in\mathbb{R}_{\geq0}^{d_A}$ over $A$'s actions can then be calculated analytically. Indeed, it is given by
\begin{align}\label{eq:transparent}
w_A
&=(1-p_{P,A})v_A
+p_{P,A} (1-p_{P,B}) M_A v_B 
+ p_{P,A} p_{P,B} (1-p_{P,A}) M_A M_B v_A \nonumber\\
&\quad+ p_{P,A} p_{P,B} p_{P,A} (1-p_{P,B}) M_A M_B M_A v_B 
+ \ldots \nonumber\\
&= (1-p_{P,A})\sum_{k=0}^\infty\left(p_{P,A} p_{P,B} M_A M_B\right)^k v_A
\nonumber\\&\quad
 + p_{P,A} (1-p_{P,B}) \sum_{k=0}^\infty\left(p_{P,B} p_{P,A} M_A M_B\right)^k M_A v_B \nonumber\\
&=\left(1-p_{P,A} p_{P,B} M_A M_B\right)^{-1}
\left((1-p_{P,A}) v_A +p_{P,A} (1-p_{P,B}) M_A v_B\right)\ ,
\end{align}
and analogously for player $B$.

Note that while both players use the opponent's predicted action against themselves to chose their action, the actual action probabilities $w_A$ and $w_B$ of the players are not correlated.

There is an analogy between decision-making with opponent transparency, leading to the two players recursively predicting each other, and iterated games, to which LOLA and SOS have been applied in Refs.~\citet{foerster18} and \citet{letcher18}, respectively.
In both cases, a player's policy needs to define how the player should respond the the ``previous'' action of the opponent. This analogy was already explored in \citet{oesterheld19}.
In Appendix~\ref{app:iterated}, we highlight some differences between the two scenarios, and investigate what the analogous results of those in the following Sec.~\ref{sec:dilemmas} look like for iterated games.

\section{Social dilemmas with opponent transparency}\label{sec:dilemmas}

In this section, we bring together the techniques introduced in Sec.~\ref{sec:lola} and Sec.~\ref{sec:simulation}, and apply them to two well-known social dilemmas, the PD and the game of chicken. 

\subsection{Preliminaries}

\subsubsection{Social dilemmas}

The available actions in the PD are cooperate (C) and defect (D). In the game of chicken they are swerving or going straight, which we will identify with cooperating and defecting, respectively, for easier comparison with the PD. Both games are described by four different payoffs, $P$ (punishment for both players defecting), $R$ (reward for both players cooperating), $T$ (temptation of defecting against a cooperator), and $S$ (sucker's payoff for cooperating with a defector). The PD and the game of chicken share the following properties: $R>P$ (mutual cooperation is preferable to mutual defection); $R>S$ (mutual cooperation is preferable to being exploited); and $T>R$ (there is an incentive to defect against a cooperator). It is often also assumed that $R>(T+S)/2$, such that mutual cooperation is preferable to alternating rounds of defection. The only difference between the two games is whether $P$ or $S$ is larger, that is, which action is preferable against a defector. In the PD, the best response against a defector is to defect oneself, while in the game of chicken, the best response is to cooperate. This makes defection a dominant strategy in the PD, while no dominant strategy is available in the game of chicken.

Tables~\ref{tab:pd} and \ref{tab:chicken} show the numerical payoffs we will use in the following.
We choose payoff differences which are sufficiently larger than unity to ensure that the parameter-to-payoff landscape is sufficiently curved, and the effects of taking into account e.g.\ how the opponent's next gradient step will affect one's own gradient become relevant. We note that in iterated games such as in \citet{foerster18} or \citet{letcher18}, a similar effect is achieved by adding up all the (discounted) payoffs of an iterated game.

\begin{table}
    \centering
    \begin{tabular}{c|c|c}
	    & $C$ (cooperate) & $D$ (defect) \\ \hline
	    $C$ (cooperate) & 30, 30 & 0, 40  \\ \hline 
	    $D$ (defect) & 40, 0  &  10, 10  
    \end{tabular}
    \caption{Payoff matrix for the prisoner's dilemma.}
    \label{tab:pd}
\end{table}

\begin{table}
    \centering
    \begin{tabular}{c|c|c}
	    & $C$ (swerve) & $D$ (straight) \\ \hline
	    $C$ (swerve) & 30, 30 & 0, 40  \\ \hline 
	    $D$ (straight) & 40, 0  &  -30, -30  
    \end{tabular}
    \caption{Payoff matrix for the game of chicken.}
    \label{tab:chicken}
\end{table}

\subsubsection{Parameter initialization and updates}

\citet{foerster18} and \citet{letcher18} consider a set up in which the two learning agents repeatedly play an \emph{iterated} PD, and update their parameters after each iterated PD. There are thus two levels of iteration: an inner one (the iterated PD), during which parameters are frozen, and an outer one, in which parameter updates happen. By contrast, we consider a single level of iteration and update parameters after each single-shot PD.

At the beginning of each experiment, we initialize all parameters from $\mathcal{N}(0, \sigma^2)$ with $\sigma=0.1$. 
In order to calculate player $A$'s gradients using Eq.~(\ref{eq:lola_exact}) or Eq.~(\ref{eq:sos_exact}), we need the parameter-to-payoff function $V_A(\theta_A, \theta_B)$. This function involves several steps. First, all parameters $\theta_A$ are translated to the probabilities appearing in $(p_{P,A}, v_A, M_A)$ using a sigmoid function. 
The probability $p_{P,A}$ is clamped to the range $[0,1-\varepsilon]$.
Then we use use Eq.~(\ref{eq:transparent}) to calculate the probability of each player cooperating, which leads to the probability of the four possible outcomes $CC$, $CD$, $DC$, and $DD$.  Multiplying the probabilities of the four possible outcomes with the respective payoffs leads to the expected payoffs $V_A$ and $V_B$.
We use PyTorch's \texttt{autograd} functionality to evaluate the gradients appearing in Eq.~(\ref{eq:lola_exact}) and Eq.~(\ref{eq:sos_exact}).

\subsection{Prisoner's dilemma}

\subsubsection{Cooperativeness as a function of the opponent learning rate}

\begin{figure}\centering
\includegraphics[width=1.0\textwidth]{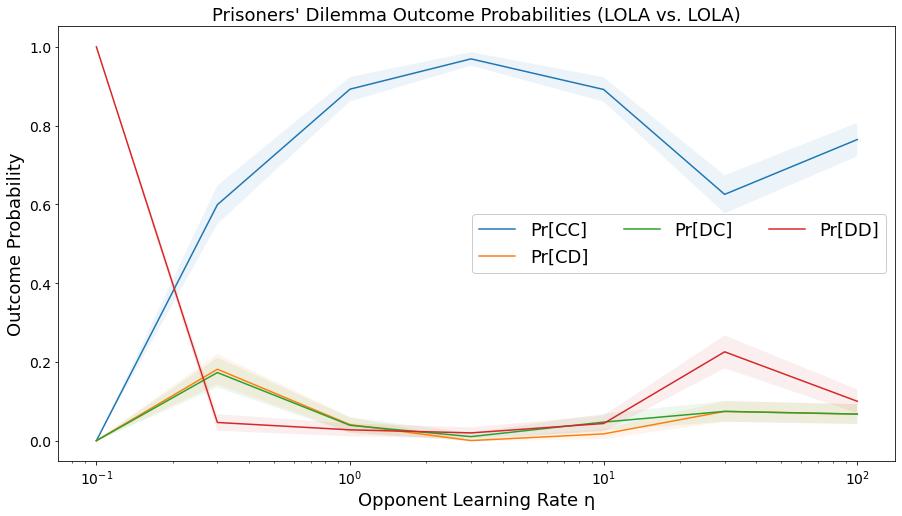}
\caption{Probability of the four different outcomes in the PD after $1000$ gradient steps of two LOLA learners as a function of $\eta=\eta_A=\eta_B$, using $\delta_A=\delta_B=1$. 
The horizontal axis is logarithmic and spans three orders of magnitude for $\eta$.
Shaded regions show one standard error calculated as $2\sigma/\sqrt{n_\text{sample}}$, where $\sigma$ is the sample standard deviation and results are sampled over $n_\text{sample}=100$ experiments.}
\label{fig:pd_by_eta}
\end{figure}

We start by studying the impact of the learning rate $\eta=\eta_A=\eta_B$ used to take the opponent's gradient step into account, while holding the actual learning rates $\delta=\delta_A=\delta_B=1$ constant. We use values $\eta\in\lbrace0.1,0.3,1,3,10,30,100\rbrace$, each learner performs $1000$ gradient steps after random initialization, and we perform $n_\text{sample}=100$ experiments for each value of $\eta$. Results for two LOLA learners are shown in Fig.~\ref{fig:pd_by_eta} and look substantially the same for SOS vs.\ SOS and LOLA vs.\ SOS.
With $\eta\ll\delta$, the LOLA learners are essentially naive learners who end up always defecting. The probability of mutual cooperation then increases with increasing $\eta$ up to a sweet spot around $\eta\approx3\delta$ where mutual cooperation is almost guaranteed. Increasing $\eta$ even further to $\eta\gg\delta$ again increases the probability of other outcomes while mutual cooperation remains the most likely outcome.

These results are quite similar to the analogous results in the iterated PD (Fig.~\ref{fig:iter_pd} in Appendix~\ref{app:iterated}).
In both cases, opponent-aware learning leads learners to take into account that making a tit-for-tat-like offer to their opponent creates an incentive for the opponent to start cooperating, without allowing a high rate of exploitation from the opponent.
\subsubsection{Fluctuations around equilibrium}

\begin{figure}\centering
\includegraphics[width=1.0\textwidth]{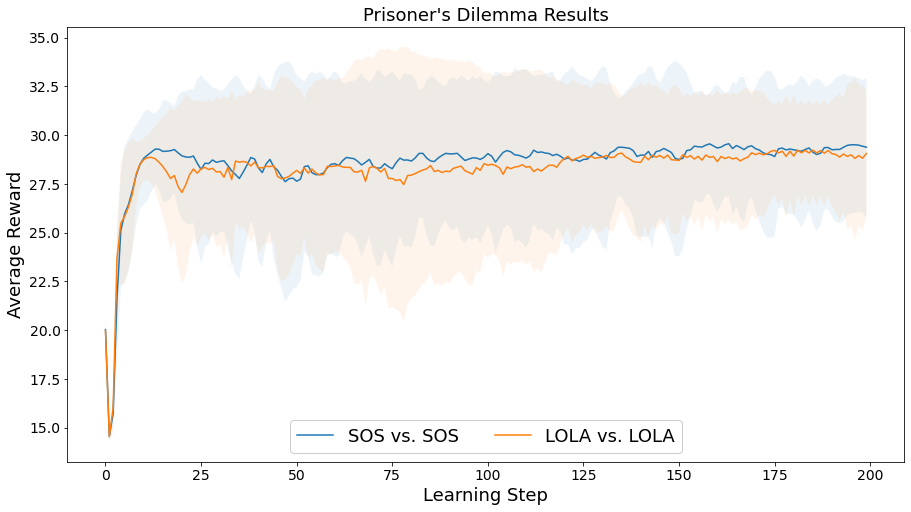}
\caption{Average payoff in a PD as a function of learning step for LOLA vs.\ LOLA learners and SOS vs.\ SOS learners, using $\delta_A=\delta_B=\eta_A=\eta_B=1$. 
Shaded regions show one standard deviation. Results are sampled over $100$ experiments.}
\label{fig:pd_fluctuations}
\end{figure}

We next study the fluctuations around the learning equilibrium in which players mostly cooperate. Fig.~\ref{fig:pd_fluctuations} shows the average payoffs and their standard deviations for LOLA vs.\ LOLA and SOS vs.\ SOS learners using $\delta_A=\delta_B=\eta_A=\eta_B=1$. Both types of learners achieve average payoffs close to the Pareto optimal value of $R=30$. At the same time, both show significant fluctuations around their average rewards after $200$ learning steps, indicating that the learning dynamics have not reached an equilibrium and learners are still trying to take advantage of each other.

\subsubsection{Parameters after convergence}\label{sec:final_params}

In the two-player, two-action games we are investigating, agents' policies are described by the four probabilities 
$\Pr\lbrack\text{S}\rbrack$ (probability of choosing an action based on predicting (``simulating'') the opponent),
$\Pr[\text{C}|\neg\text{S}]$ (probability of cooperating when giving an opponent-independent response), 
$\Pr[\text{C}|\text{C}]$ (probability of cooperating after opponent's simulated cooperation), and
$\Pr[\text{C}|\text{D}]$ (probability of cooperating after opponent's simulated defection).
We will now discuss the values towards which these probabilities converge.

\begin{figure}\centering
\makebox[\textwidth]{
\noindent\begin{tabular}{c}
      \includegraphics[width=\textwidth]{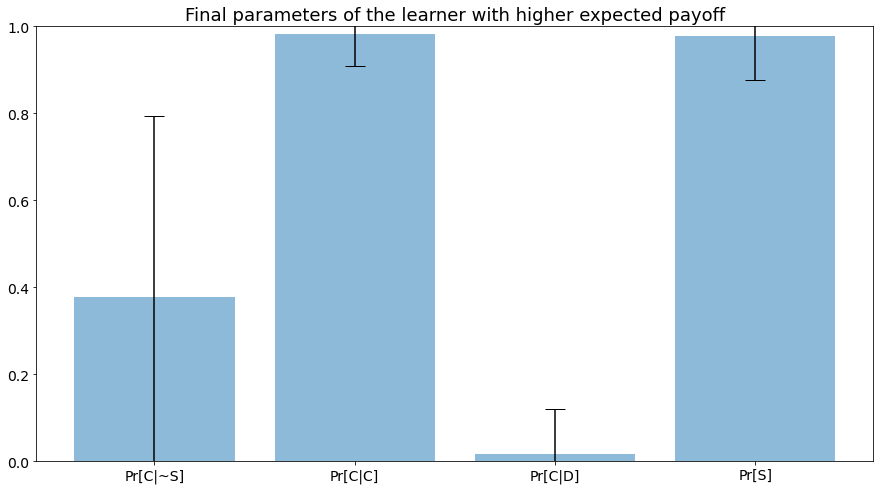} \nonumber\\
       \includegraphics[width=\textwidth]{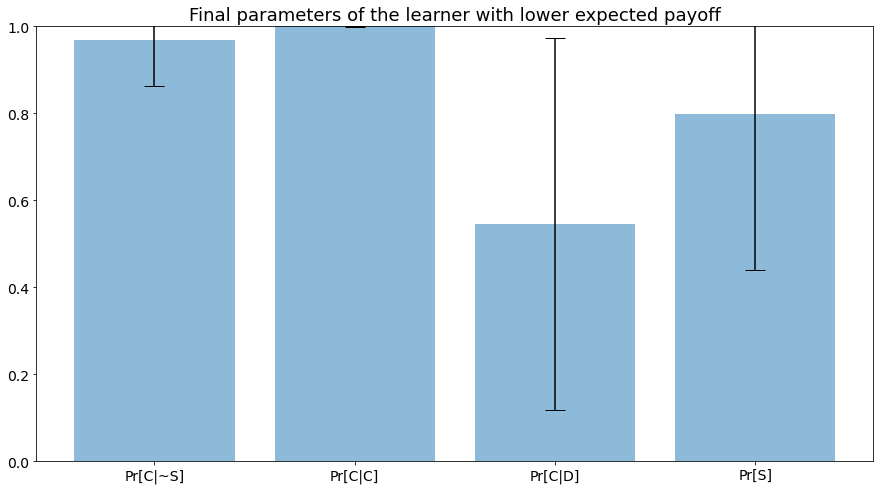}
\end{tabular}
}
\caption{Final parameters in a PD of LOLA learners with $\delta_A=\delta_B=\eta_A=\eta_B=1$ after $1000$ learning steps.
Parameters of the learner who achieves higher expected payoff are at the top. Error bars show one standard deviation. 
}
\label{fig:pd_params}
\end{figure}

Learners with $\delta_A=\delta_B=\eta_A=\eta_B=1$ achieve final payoffs (after $1000$ learning steps) which are close to identical: the higher payoff is $29.19 \pm 2.14$ (standard deviation over $100$ experiments), the lower payoff is $28.74 \pm 3.19$.
Interestingly, the final parameters are clearly not identical, see Fig.~\ref{fig:pd_params}.
It is instructive to consider how these final parameters differ from $\varepsilon\texttt{GroundedFairBot}$ ($\Pr\lbrack\text{S}\rbrack=1-\varepsilon$, $\Pr[\text{C}|\text{C}]=\Pr[\text{C}|\neg\text{S}]=1$, $\Pr[\text{C}|\text{D}]=0$).
Both learners converge to $\Pr[\text{C}|\text{C}]\approx1$, which provides an incentive for their opponent to cooperate. The learner with higher final payoff differs from $\varepsilon\texttt{GroundedFairBot}$ by defecting with significant probability when giving an opponent-independent response, the learner with lower final payoff differs from $\varepsilon\texttt{GroundedFairBot}$ by choosing an opponent-independent action with significant probability, and cooperating after simulated defection with significant probability.

We can provide a heuristic explanation how such an equilibrium can emerge.
When playing against ``almost $\varepsilon\texttt{GroundedFairBot}$'' (say an agent that plays like $\varepsilon\texttt{GroundedFairBot}$ $99\%$ of the time, and randomly $1\%$ of the time), it is better to cooperate unconditionally than to be $\varepsilon\texttt{GroundedFairBot}$.
This can be seen with the help of Eq.~(\ref{eq:transparent}), or by considering that cooperating unconditionally prevents (simulated and real) retribution.
Starting from a hypothetical state in which both agents are ``almost $\varepsilon\texttt{GroundedFairBot}$'', there is thus a gradient towards cooperating unconditionally, i.e.\ increasing $\Pr[\text{C}|\text{D}]$ and lowering $\Pr\lbrack\text{S}\rbrack$.
If $A$ starts moving towards unconditional cooperation, this invites defection from $B$, i.e.\ $B$ is invited to lower their $\Pr[\text{C}|\neg\text{S}]$. The opponent-shaping gradient of $A$ will thus push against moving too far away from $\varepsilon\texttt{GroundedFairBot}$.
In effect, in the rare cases where $B$ decides to act without simulating $A$, they have a chance to defect against a cooperating opponent, which yields the highest possible reward $T$; this will lead to a slightly higher expected payoff for $B$ than for $A$.
Which of the two learners ends in which of the roles just described will be determined by small differences in the initial parameters.

\subsubsection{The need for opponent-aware learning}

We know that the parameter space available to the learners includes policies like $\varepsilon\texttt{GroundedFairBot}$, which provide incentives for the opponent to cooperate while being hard to exploit.
We have seen in Fig.~\ref{fig:pd_by_eta} that naive learners (or LOLA learners with $\eta\ll\delta$) do not converge towards such policies, but learn mutually defecting strategies.
Here, we investigate whether mutual cooperation is at least stable for naive learners when they are initialized with parameters close to $\varepsilon\texttt{GroundedFairBot}$.
We initialize all parameters to $\theta=\pm3$ plus small noise from $\mathcal{N}(0, \sigma^2)$ with $\sigma=0.1$, 
such that initially $\Pr\lbrack\text{S}\rbrack\approx\Pr[\text{C}|\neg\text{S}]\approx\Pr[\text{C}|\text{C}]\approx95\%$ and $\Pr[\text{C}|\text{D}]\approx5\%$.


\begin{figure}\centering
\includegraphics[width=0.7\textwidth]{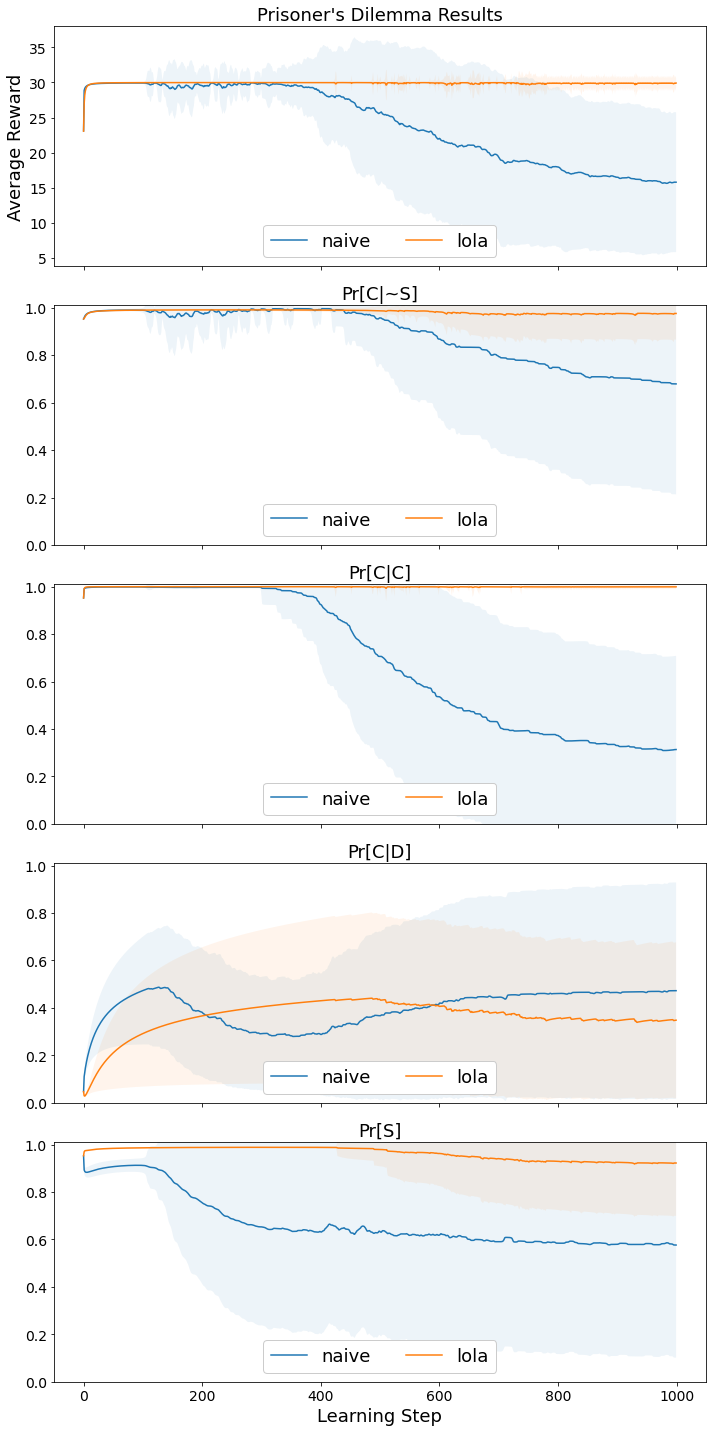}
\caption{Prisoner's dilemma with learners whose parameters are initialized close to $\varepsilon\texttt{GroundedFairBot}$ ($\Pr\lbrack\text{S}\rbrack\approx\Pr[\text{C}|\neg\text{S}]\approx\Pr[\text{C}|\text{C}]\approx95\%$ and $\Pr[\text{C}|\text{D}]\approx5\%$).
Both opponents use the same learning algorithm.
They are either naive gradient learners (with $\delta_A=\delta_B=1$) or LOLA learners (with $\delta_A=\delta_B=\eta_A=\eta_B=1$).
Shaded regions show one standard deviation sampled over $100$ experiments.
}
\label{fig:pd_with_without_lola}
\end{figure}

Fig.~\ref{fig:pd_with_without_lola} compares the outcomes with and without opponent-aware learning. Both two naive learners and two LOLA learners initially achieve high levels of cooperation. However, for the former cooperation starts degrading after a few hundred learning steps, while for the latter it remains stable over $1000$ steps.
As already discussed in Sec.~\ref{sec:final_params}, the reason for this is that when playing against ``almost $\varepsilon\texttt{GroundedFairBot}$'', the naive gradient points towards cooperating unconditionally, i.e.\ lowering $\Pr\lbrack\text{S}\rbrack$. This can be observed for the naive learners in  Fig.~\ref{fig:pd_with_without_lola}. 
As a consequence, the incentive to reward simulated cooperation with cooperation decreases, leading to decreasing $\Pr[\text{C}|\text{C}]$.
When the opponent-shaping gradient is present, $\Pr\lbrack\text{S}\rbrack$ stays close to unity, as seen for the LOLA learners.

Interestingly, both naive and LOLA learners end up more forgiving than $\varepsilon\texttt{GroundedFairBot}$ (i.e., both have $\Pr[\text{C}|\text{D}] > 0$). LOLA learners manage to strike a balance between forgiving simulated defection with significant probability (thus preventing cascades of simulated and real defection) and keeping $\Pr[\text{C}|\text{D}]$ low enough such that both agents are still incentivized to cooperate with high probability when choosing an opponent-independent action ($\Pr[\text{C}|\neg\text{S}]\approx1$).

\subsubsection{No arms race towards higher opponent learning rates}

\begin{figure}\centering
\includegraphics[width=1.0\textwidth]{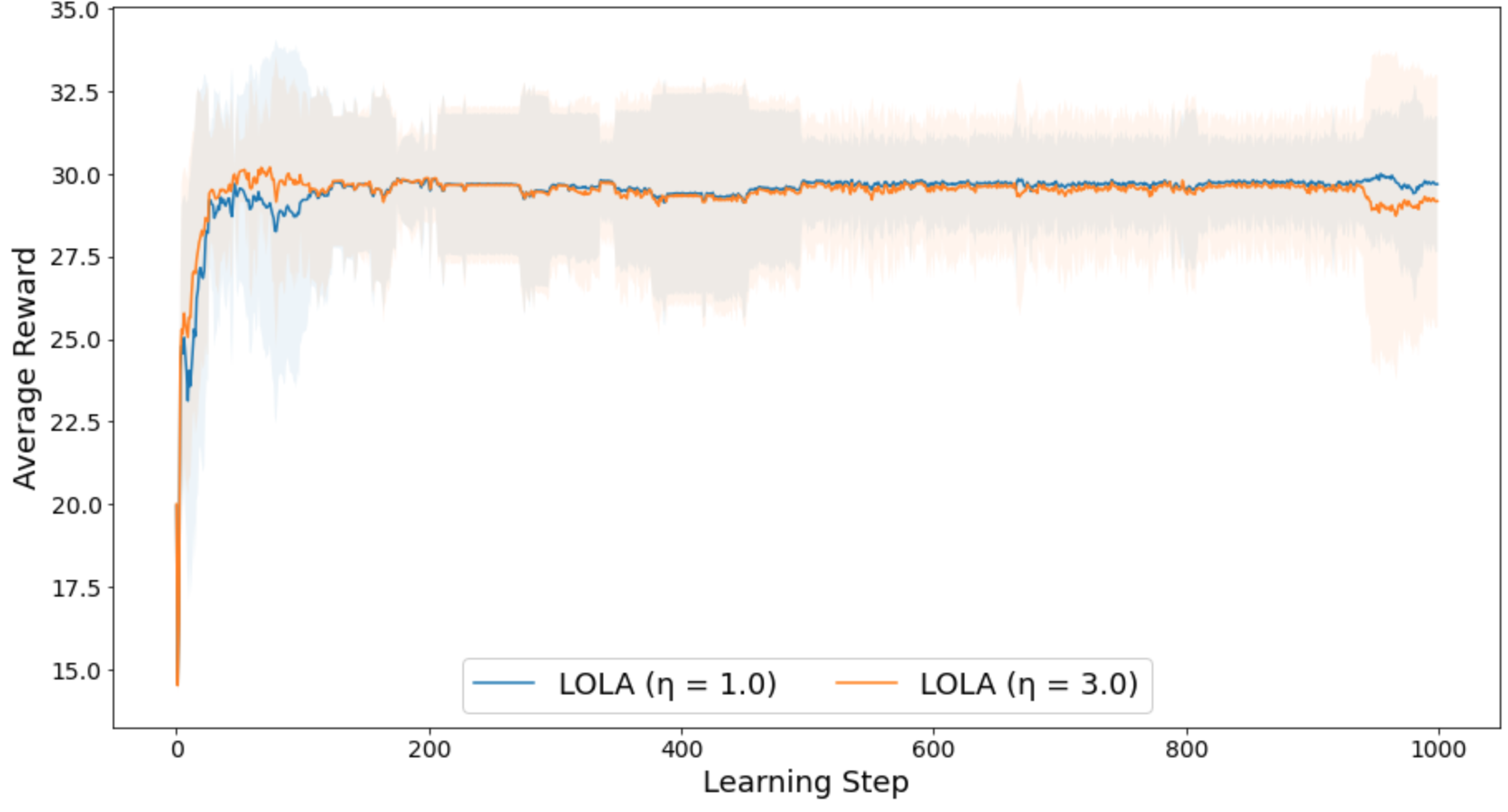}
\includegraphics[width=1.0\textwidth]{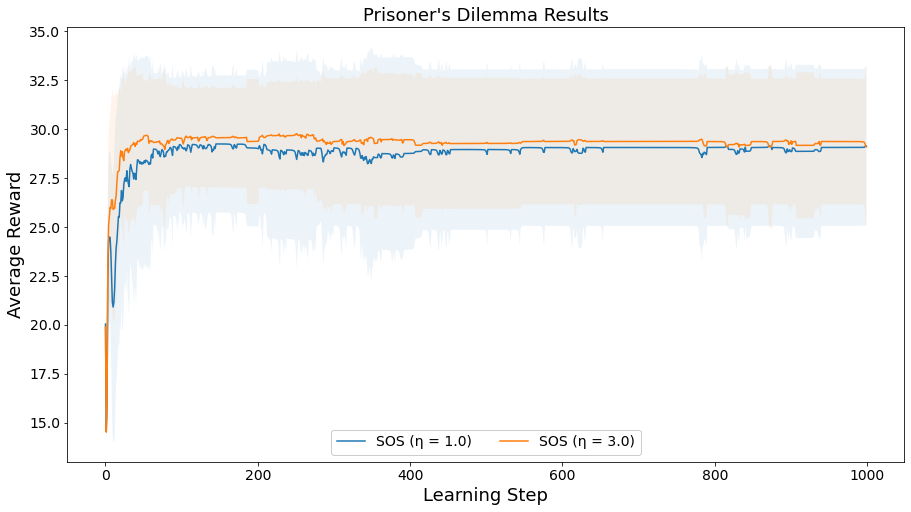}
\caption{Average payoff in a PD as a function of learning step for LOLA vs.\ LOLA learners (top) and SOS vs.\ SOS learners (bottom), using $\delta_A=\delta_B=\eta_A=1$ and $\eta_B=3$. 
Shown are average payoffs for the first $1000$ learning steps.
Shaded regions show one standard deviation. Results are sampled over $100$ experiments.}
\label{fig:pd_unequal}
\end{figure}

In a competitive scenario there is no reason to assume that both learners use the same opponent learning rate $\eta$. 
Indeed, in a scenario in which principals submit learning agents to compete with others, the principals might endow their agents with a high $\eta$ in the hope of their agent outwitting its opponent. It is thus interesting to study whether a learner with higher $\eta$ has an advantage over a learner with lower $\eta$ (which might lead to an ``arms race'' of learners with ever higher $\eta$), and how stable mutual cooperation is to learners with unequal opponent learning rates.
Fig.~\ref{fig:pd_unequal} shows results for $\delta_A=\delta_B=\eta_A=1$ and $\eta_B=3$. The learner using the higher opponent learning rate $\eta_B=3$ does not manage to achieve higher payoffs on average. For both two LOLA and two SOS learners, cooperation is stable over $1000$ learning steps. 

\subsubsection{The game from the perspective of the principals}

We now take the scenario of two principals submitting learning agents one step further. We can consider the ``game'' in which a strategy does not correspond to cooperating with some probability (as in the underlying PD), but to submitting a certain learning agent that will play multiple rounds of the underlying game (like the PD) against its opponent while updating its parameters. The principal then receives the reward of a final single-shot game. 
(For sufficiently long training periods, results do not change substantially if the principals instead receive the sum of all rewards received by their agent during training.) 
In such a scenario, each principal might be concerned not only with the learning of their own agent, but also with how they can shape the learning of their opponent's agent.

Within the scope of this work, a learning agent is defined by the algorithm used to calculate its gradients (naive, LOLA, or SOS), its learning rate $\delta$, and its opponent learning rate $\eta$ (for non-naive learners). This still produces a large space of possible learning agents. For simplicity, we fix $\delta=1$ for all learners and evaluate the impact of different values of $\eta$ for both LOLA and SOS.

\begin{figure}\centering
\includegraphics[width=1.0\textwidth]{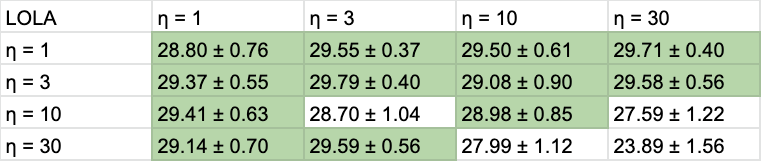}
\includegraphics[width=1.0\textwidth]{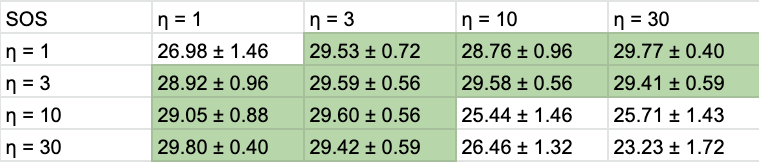}
\caption{Average payoff in the final single-shot PD after $1000$ learning steps. Shown is the payoff of the row strategy versus the column strategy (top: LOLA vs.\ LOLA; bottom: SOS vs.\ SOS). All strategies use a learning rate $\delta=1$. Error bars are calculated as $2\sigma/\sqrt{n_\text{sample}}$, where $\sigma$ is the sample standard deviation and $n_\text{sample}=100$ is the number of experiments per strategy pair. Cells are highlighted if the row strategy is a best response versus the column strategy within statistical error bars.}
\label{fig:pd_learners}
\end{figure}

Fig.~\ref{fig:pd_learners} shows the results (average reward received in the final single-shot game). We find that intermediate opponent learning rates ($\delta\approx3$) are close to dominant, i.e.\ produce close to the optimal payoff against all opponents. When facing each other, strategies with $\eta=3$ produce payoffs close to the Pareto-optimal value $R=30$. The framework of learning with opponent transparency thus transforms the PD, with its default outcome of mutual defection, into a much easier game in which mutual cooperation is the default outcome.

\subsection{Game of chicken}

\subsubsection{Baseline outcome for naive learners}

Two naive learners in the PD simply learn to always defect against each other, even when given the opportunity to predict their opponent against themselves.
In the game of chicken, defection is no longer the best response against a defecting opponent, and so mutual defection is not a stable outcome for naive learners. Indeed, one of the two naive learners will learn to always defect while the other learns to always cooperate, yielding an average payoff of $(S+T)/2 < R$. Which of the two learners becomes the ``defector'' and which one the ``cooperator'' is determined by small differences in the initial parameters.



\subsubsection{Outwitting the opponent}

\begin{figure}\centering
\makebox[\textwidth]{
\noindent\begin{tabular}{cc}
      \includegraphics[width=0.5\textwidth]{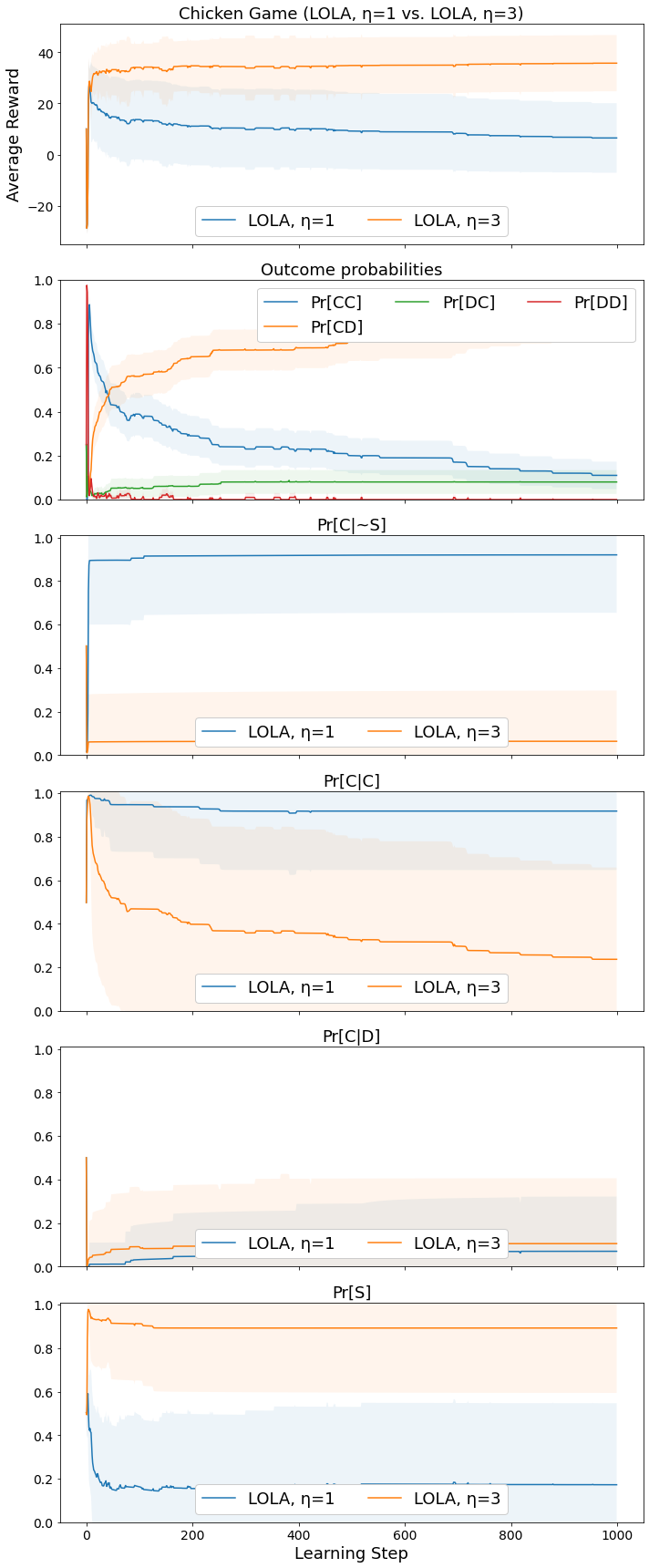} &
      \includegraphics[width=0.5\textwidth]{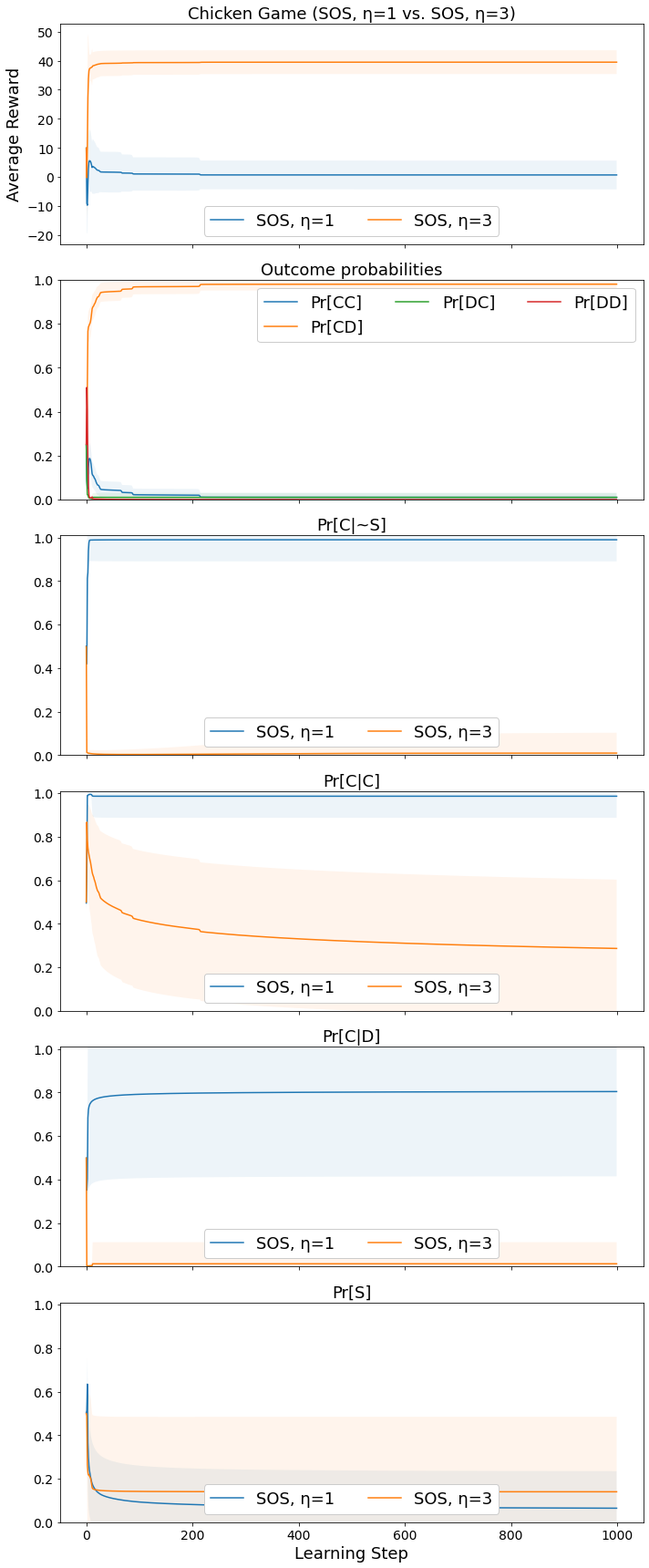}
\end{tabular}
}
\caption{Two examples of a learner outmaneuvering their opponent in a game of chicken.
Left: Two LOLA learners with $\delta_A=\delta_B=\eta_A=1$ and $\eta_B=3$.
Right: Two SOS learners with $\delta_A=\delta_B=\eta_A=1$ and $\eta_B=3$.
In the ``Outcome probabilities'' plot, shaded regions show one standard error calculated as $2\sigma/\sqrt{n_\text{sample}}$, where $\sigma$ is the sample standard deviation; in the other plots, shaded regions show one sample standard deviation.
Results are sampled from $n_\text{sample}=100$ experiments.
}
\label{fig:chk_different_learners}
\end{figure}

When using opponent-aware learners instead of naive learners, it becomes possible for one learner to navigate its opponent towards its preferred equilibrium.
Fig.~\ref{fig:chk_different_learners} shows two examples of this. In both of these examples, one learner manages to get its opponent to cooperate with near-certainty, while itself mostly defecting. As a result, it ends up receiving an average reward close to $T=40$ at test time.
The left part of the figure shows a LOLA learner with higher opponent learning rate $\eta$ taking advantage of its opponent with lower $\eta$. The right part shows the same for two SOS learners.

In the game of chicken, one promising opponent-aware strategy might be to move towards unconditional defection ($\Pr[\text{S}]\rightarrow0$ and $\Pr[\text{C}|\neg\text{S}]\rightarrow0$), forcing the opponent to cooperate lest they receive the lowest possible payoff $P$. 
(This strategy in the game of chicken is sometimes described as ``throwing the steering wheel out the window'', i.e.\ visibly committing to going straight.)
We observe from  Fig.~\ref{fig:chk_different_learners} that this is indeed the strategy followed by the higher-$\eta$ SOS learner.

\subsubsection{Outcomes as a function of the opponent learning rate}

\begin{figure}\centering
\includegraphics[width=0.75\textwidth]{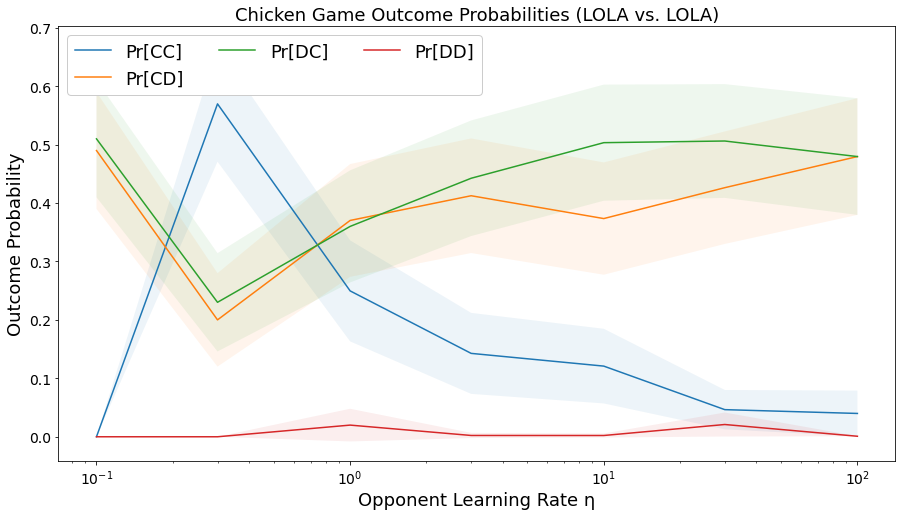} \nonumber\\
\includegraphics[width=0.75\textwidth]{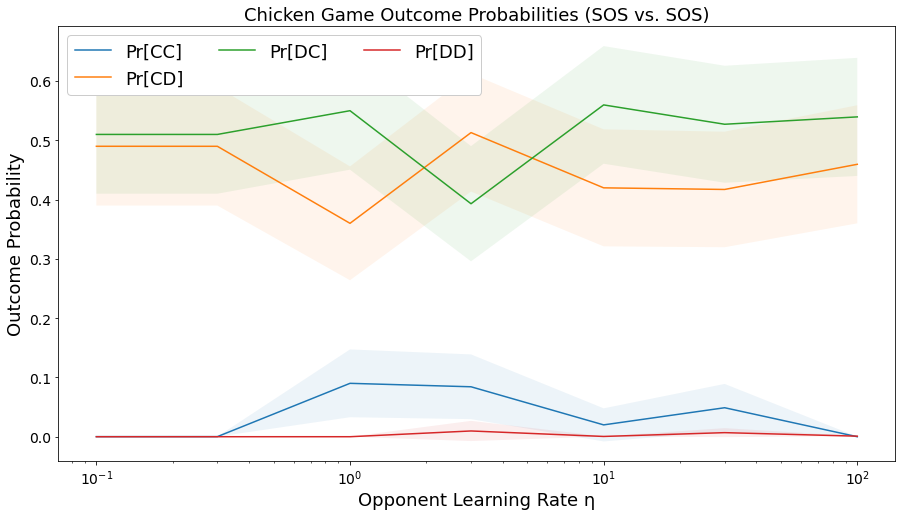} 
\caption{Outcome probabilities (after $1000$ gradient steps) in a game of chicken as a function of $\eta_A=\eta_B=\eta$ (while holding $\delta_A=\delta_B=1$ constant).
The top figure shows LOLA vs.\ LOLA, the bottom figure SOS vs.\ SOS.
The horizontal axis is logarithmic and spans three orders of magnitude for $\eta$.
Shaded regions show one standard error calculated as $2\sigma/\sqrt{n_\text{sample}}$, where $\sigma$ is the sample standard deviation and $n_\text{sample}=100$.}
\label{fig:chk_by_eta}
\end{figure}

Fig.~\ref{fig:chk_by_eta} shows the results of both learners using ever larger values of $\eta_A=\eta_B=\eta$ (while holding $\delta_A=\delta_B=1$ constant).
For sufficiently small values of $\eta$, one of the players will always cooperate while their opponent will always defect. 
For two LOLA learners (top), there are intermediate values of $\eta$ for which mutual cooperation becomes the most likely outcome.
For two SOS learners (bottom), the outcome probabilities never differ significantly from the $\eta=0$  baseline (C/D and D/C with equal probability).

\begin{figure}\centering
\includegraphics[width=0.75\textwidth]{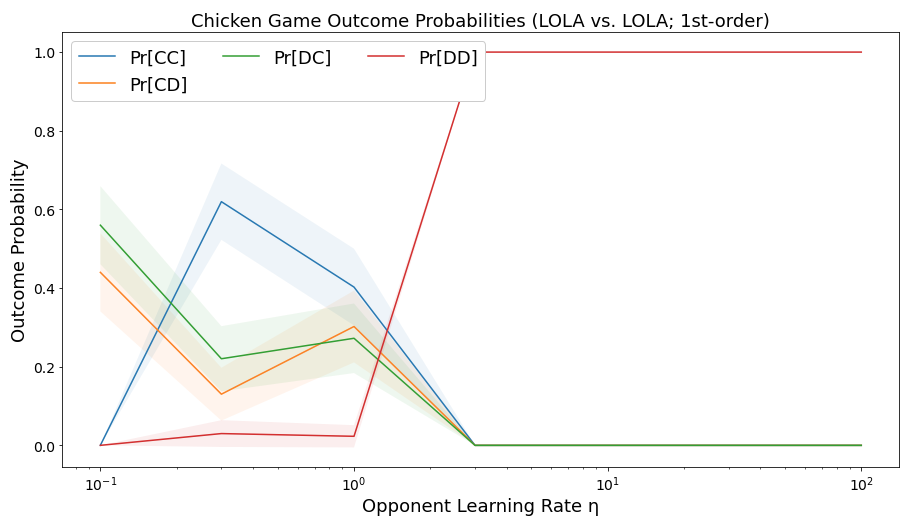} \nonumber\\
\includegraphics[width=0.75\textwidth]{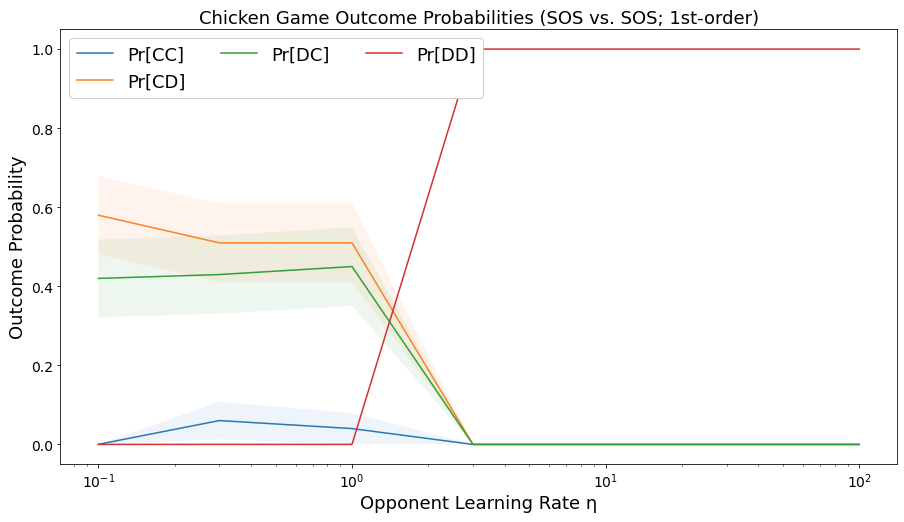}
\caption{Outcome probabilities (after $1000$ gradient steps) in a game of chicken as a function of $\eta_A=\eta_B=\eta$ (while holding $\delta_A=\delta_B=1$ constant).
The top figure shows LOLA vs.\ LOLA using the first-order approximation Eq.~(\ref{eq:lola_1st_order}) to LOLA, the bottom figure shows SOS vs.\ SOS using the first-order approximation Eq.~(\ref{eq:sos_1st_order}) to SOS.
The horizontal axis is logarithmic and spans three orders of magnitude for $\eta$.
Shaded regions show one standard error calculated as $2\sigma/\sqrt{n_\text{sample}}$, where $\sigma$ is the sample standard deviation and $n_\text{sample}=100$.}
\label{fig:chk_by_eta_1st}
\end{figure}

So far we considered the ``exact'' version of LOLA and SOS, i.e. Eqs.~(\ref{eq:lola_exact}) and (\ref{eq:sos_exact}), respectively.
However, when these algorithms were originally introduced in \citet{foerster18} and \citet{letcher18}, respectively, 
their first-order versions Eqs.~(\ref{eq:lola_1st_order}) and (\ref{eq:sos_1st_order}) were proposed.
We find that the outcome probabilities in the game of chicken change drastically (for large $\eta$) when using these first-order versions, see Fig.~\ref{fig:chk_by_eta_1st}.

When using these first-order approximations and large enough $\eta$, LOLA and SOS learners display ``arrogant behavior'' in that they expect that them defecting will compel their opponent to cooperate. This is especially noteworthy since SOS was specifically designed to avoid such arrogant behavior. The reason why SOS does not manage to avoid arrogant behavior here is that the SOS gradient $\xi_{\text{SOS},A}$ is designed to have non-negative inner product with the Look Ahead gradient $\xi_{\text{LA},A}$ (see Appendix~\ref{app:sos}), not necessarily with the naive gradient $\xi_{\text{naive},A}$. If both players are defecting, the naive gradient will point towards defecting less, but the Look Ahead gradient might (for large enough $\eta$) not, since it anticipates the opponent defecting less. 
This provides a clear example in which incorrectly anticipating one's opponent proves harmful.

\subsubsection{The game from the perspective of the principals}

\begin{figure}\centering
\includegraphics[width=1.0\textwidth]{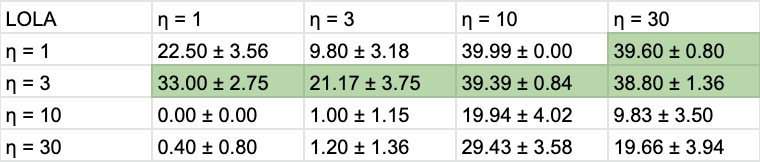}
\includegraphics[width=1.0\textwidth]{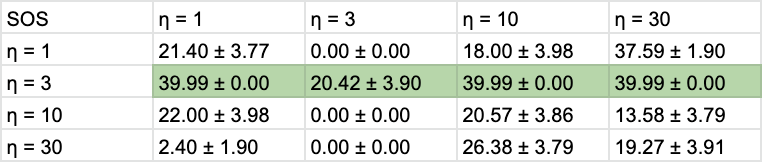}
\caption{Average payoff in the final single-shot game of chicken after $1000$ learning steps. Shown is the payoff of the row strategy versus the column strategy (top: LOLA vs.\ LOLA; bottom: SOS vs.\ SOS). All strategies use a learning rate $\delta=1$. Error bars are calculated as $2\sigma/\sqrt{n_\text{sample}}$, where $\sigma$ is the sample standard deviation and $n_\text{sample}=100$ is the number of experiments per strategy pair. Cells are highlighted if the row strategy is a best response versus the column strategy within statistical error bars.}
\label{fig:chk_learners}
\end{figure}

We now consider again the scenario in which two principals submit learning agents that will play on their behalf.
From Fig.~\ref{fig:chk_learners}, we see that -- much more so than was the case in the PD -- an intermediate opponent learning rate of $\eta=3$ is dominant and able to take advantage of opponents with both higher and lower opponent learning rates. When facing each other, learning agents with $\eta=3$ get payoffs close to the baseline of $(S+T)/2=20$. The framework of learning with opponent transparency thus provides little benefit in the game of chicken.

\section{Discussion}\label{sec:discussion}

We have studied a scenario in which two principals submit learning agents that repeatedly interact in a social dilemma. At test time, the agents play a single round of the social dilemma game, and the resulting rewards are the principals' payoffs. Both players' decision making is mutually transparent, and both learners are aware of each others' learning, and take it into account when updating their own parameters. It is not at all clear how to do this ``optimally''. 
We have considered two approaches discussed in previous literature, LOLA and SOS, after adapting them to our set up. We have seen that LOLA and SOS learners can achieve outcomes which are drastically different than those achieved by naive gradient descent learners. In particular, they can achieve mutual cooperation in the  prisoner's dilemma, yielding Pareto optimal outcomes, but also (when combining a leading-order approximation with high opponent learning rates) mutual defection in the game of chicken, which yields the worst possible outcome for both players.
The latter point illustrates that opponent-awareness can be harmful in situations in which both parties model the opponent as making the ``sensible'' choice of moving out of the way.

LOLA and SOS have an additional free parameter compared with vanilla gradient learners, 
namely the opponent learning rate which describes how far the opponent's updates are anticipated.
Intuitively, there is a trade-off involved when choosing this opponent learning rate: 
When choosing it too low, opponent-aware learning doesn't provide any benefits relative to vanilla gradient learning; 
when choosing it too high, the assumption that the opponent will update its parameters along the directions dictated by current gradients becomes bad;
so we expect intermediate opponent learning rates to be optimal.
Indeed, we have found that intermediate learning rates achieve the best performance, and (in the game of chicken) are able to exploit opponents with both higher and lower opponent learning rates.

From the principals' perspective, our framework of learning with opponent transparency transforms the underlying social dilemma game into a different game: what learning agent is a good response to a learning agent which their opponent might submit? We have seen that the prisoner's dilemma and the game of chicken behave very differently under this transformation. In the prisoner's dilemma, there are learning agents which are a best response against themselves (within the limited set of learning agents considered in this work), and lead to mutual cooperation with high probability. 
In the game of chicken, by contrast, transparency seems to provide little help in overcoming the fundamental issue that both parties attempt to maneuver to their preferred equilibrium, which is different from the opponent's. 

In the prisoner's dilemma, mutual awareness of each other's inner workings is helpful for achieving cooperative outcomes. This idea has existed for decades (at least since \citet{hofstadter83}), and we have now validated it in the paradigm of gradient-based machine learning. On the other hand, we have seen that in the game of chicken, in which both learning agents attempt to navigate the opponent towards their preferred equilibrium, it is even possible that opponent-aware learners produce the worst possible outcome for themselves and so do worse than more naive learning agents would. Developing techniques that are guaranteed to achieve acceptable outcomes in games that involve an equilibrium selection problem is thus an important open problem if we want to avoid worst case outcomes in multi-agent interactions.

\section*{Acknowledgements}

Many thanks to Jakob Foerster and Alistair Letcher for answering questions about opponent-aware learning,
to Jesse Clifton and Daniel Kokotajlo for helpful discussions,
and to Caspar Oesterheld for careful reading of the manuscript.

\begin{appendix}

\section{Calculating the scaling parameter in SOS}\label{app:sos}

This appendix describes how to calculate the scaling parameter $p\in[0,1]$ in SOS. It is based on \citet{letcher18} and Remark~4.7 in \citet{letcher18Thesis} and adapted to our framework. The calculation involves two hyperparameters $a, b\in(0,1)$, where we follow the recommendation of \citet{letcher18} to use $a=0.5$ and $b=0.1$.

Let us define the Look Ahead (LA) gradient
\begin{align}
\xi_{\text{LA},A}=\xi_{\text{naive},A}+\eta_A\nabla_{\theta_B}V_B(\bar{\theta}_A,\bar{\theta}_B)\nabla_{\theta_B}\nabla_{\theta_A}V_A(\bar{\theta}_A,\bar{\theta}_B)    
\end{align}
given by the first two summands of Eq.~(\ref{eq:lola_1st_order}). Further let
\begin{align}
\chi_A = \eta_A\nabla_{\theta_B}V_A(\bar{\theta}_A,\bar{\theta}_B)    
\end{align}
denote the leading-order opponent-shaping correction, i.e.\ the third summand in Eq.~(\ref{eq:lola_1st_order}). The SOS gradient can then be written as
\begin{align}
\xi_{\text{SOS},A} = \xi_{\text{LA},A} + p\chi_A\ .
\end{align}
Let $p_1=1$ if $\langle\chi_A,\xi_{\text{LA},A}\rangle\geq0$ and $p_1=\min\{1,-a\|\xi_{\text{LA},A}\|^2/\langle\chi_A,\xi_{\text{LA},A}\rangle\}$ otherwise.
Let $p_2=\|\xi_{\text{naive},A}\|^2$ if $\|\xi_{\text{naive},A}\|<b$ and $p_2=1$ otherwise. 
Choosing $p\leq p_1$ ensures that $\xi_{\text{SOS},A}$ always has non-negative inner product with $\xi_{\text{LA},A}$.
Choosing $p\leq p_2$ ensures that SOS converges to SFPs.
In order to guarantee both of these while making the opponent-shaping correction as strong as possible, choose $p=\min\lbrace p_1,p_2\rbrace$.

\section{Comparison with iterated games}\label{app:iterated}

Both players recursively predicting each other, as formalized in Eq.~(\ref{eq:transparent}), is somewhat analogous to an iterated game with one-step memory. In both cases, players need to decide how to map the opponent's (previous or predicted) action to their response. 
This analogy was already explored in \citet{oesterheld19}.
However, it's worth highlighting a few differences:
\begin{itemize}
    \item When predicting the opponent, a policy needs to map from the space of possible opponent actions to the space of own actions. 
    By contrast, in an iterated game the mapping is from the Cartesian product of possible own actions and opponent actions to one's own actions.
    This allows for a richer set of strategies in iterated games. For example, ``Win-stay Lose-shift'' (e.g. \cite{biol07}) is a prominent strategy in the iterated PD, but has no analogous strategy when choosing one's action based on the opponent's predicted action. 
    \item In the iterated game, players receive a reward at each step, while during recursive mutual prediction, they only receive a single reward ``at the end'', when choosing their actual action.
    \item In an iterated game, both players take actions on the basis of the same history. By contrast, when both players predict each other, there is a separate recursion for both of them. In particular, in an iterated game the game has the same duration (or the same discounting in the case of an infinitely iterated game) for both players, while the recursive mutual predictions of the two players can have different lengths.
\end{itemize}

\begin{figure}\centering
\includegraphics[width=0.75\textwidth]{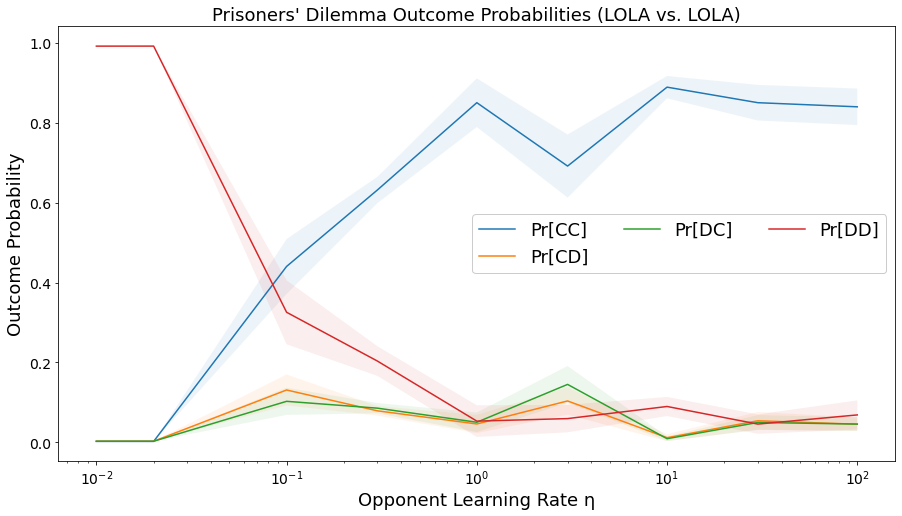} \nonumber\\
\includegraphics[width=0.75\textwidth]{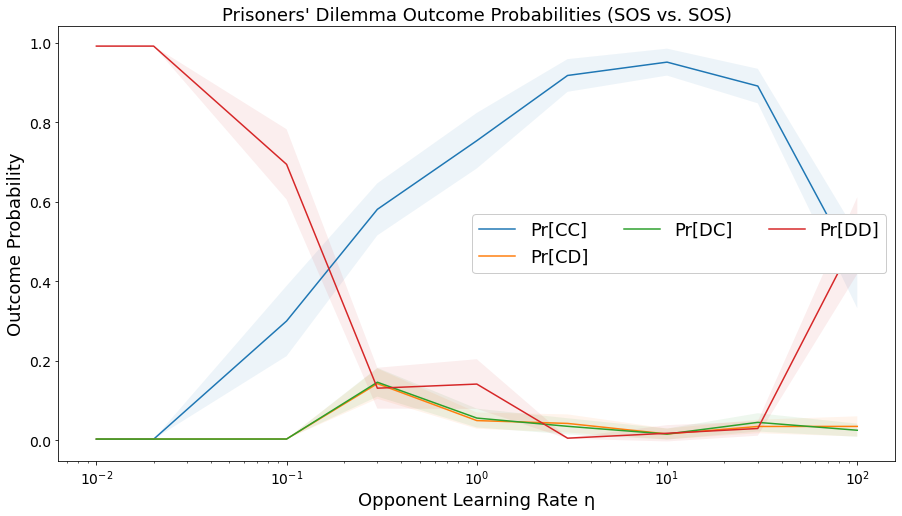} 
\caption{Outcome probabilities after $1000$ iterated PDs, as a function of $\eta_A=\eta_B=\eta$ (while holding $\delta_A=\delta_B=1$ constant).
The top figure shows LOLA vs.\ LOLA, the bottom figure SOS vs.\ SOS.
The horizontal axis is logarithmic and spans three orders of magnitude for $\eta$.
Shaded regions show one standard error calculated as $2\sigma/\sqrt{n_\text{sample}}$, where $\sigma$ is the sample standard deviation and $n_\text{sample}=100$.}
\label{fig:iter_pd}
\end{figure}

\begin{figure}\centering
\includegraphics[width=0.75\textwidth]{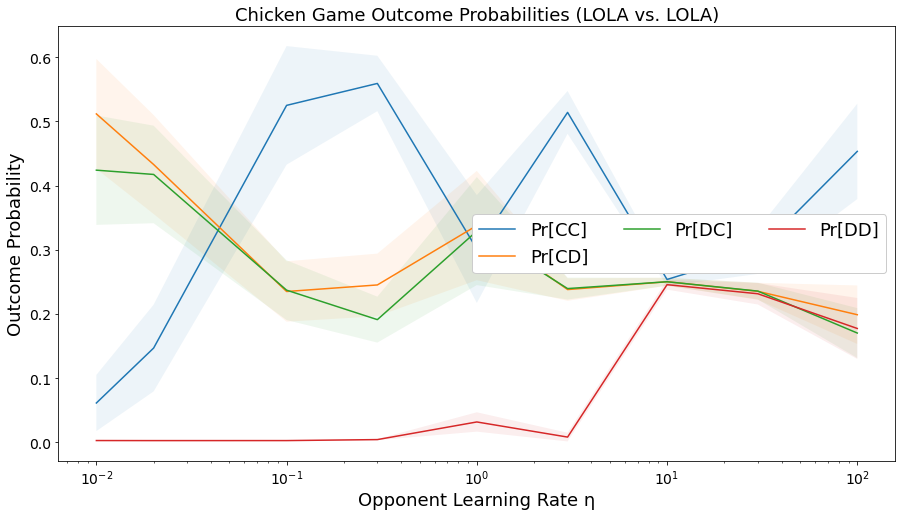} \nonumber\\
\includegraphics[width=0.75\textwidth]{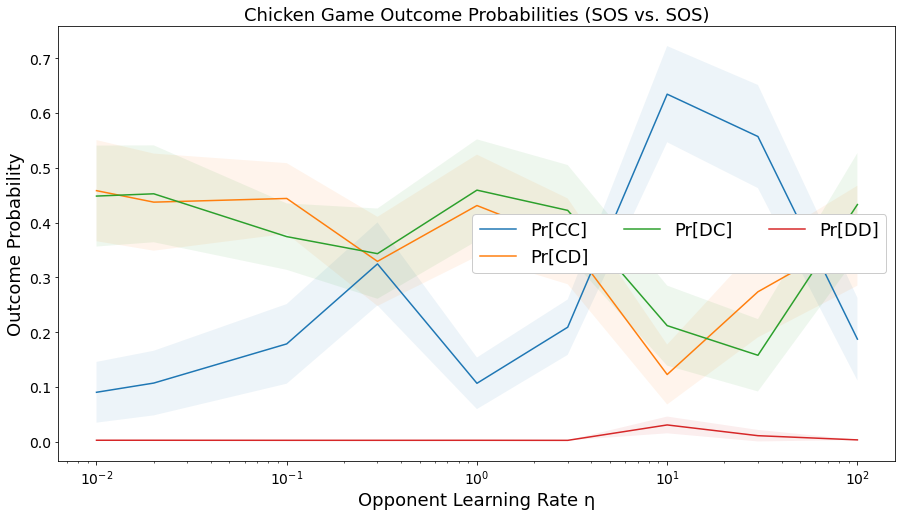} 
\caption{Outcome probabilities after $1000$ iterated games of chicken, as a function of $\eta_A=\eta_B=\eta$ (while holding $\delta_A=\delta_B=1$ constant).
The top figure shows LOLA vs.\ LOLA, the bottom figure SOS vs.\ SOS.
The horizontal axis is logarithmic and spans three orders of magnitude for $\eta$.
Shaded regions show one standard error calculated as $2\sigma/\sqrt{n_\text{sample}}$, where $\sigma$ is the sample standard deviation and $n_\text{sample}=100$.}
\label{fig:iter_chk}
\end{figure}

Given these differences, it is interesting to investigate what the analogous plots to Figs.~\ref{fig:pd_by_eta} and \ref{fig:chk_by_eta} look like in an iterated game. We consider an infinitely repeated iterated game with a discounting factor $\gamma=0.99$. The expected payoffs for given (fixed) policies can be calculated in closed form (\cite{foerster18}). We use the payoff matrices in tables~\ref{tab:pd} and \ref{tab:chicken}, and, to keep the scale of payoffs identical to the single-shot case with mutual prediction, use the averaged (discounted) reward per step for updating parameters. 
Figs.~\ref{fig:iter_pd} and \ref{fig:iter_chk} show the outcome probabilities for both iterated games as a function of the opponent learning rate $\eta$.

As in the case of single-shot games with mutual prediction, mutual cooperation (CC) becomes the most likely outcome in an iterated PD for intermediate values of $\eta$ ($\sim 1-10)$. In the game of chicken, there is a higher rate of mutual cooperation than in the single-shot case with mutual prediction for a wide range of values of $\eta$.

\end{appendix}

\bibliographystyle{apalike}
\bibliography{bibliography}

\end{document}